\documentclass[11pt]{article}

\usepackage[]{acl}
\usepackage{array} 

\usepackage{times}
\usepackage{latexsym}

\usepackage[T1]{fontenc}

\usepackage[utf8]{inputenc}

\usepackage[T1]{fontenc}

\usepackage{microtype}

\usepackage{inconsolata}

\usepackage{graphicx}
\usepackage{booktabs}
\usepackage{xcolor} 
\usepackage{multirow}
\usepackage{hhline}
\usepackage{booktabs}
\usepackage{xspace}
\usepackage{makecell}
\usepackage{vcell}
\usepackage{hyperref}
\usepackage{float}
\usepackage{subcaption}

\usepackage[table]{xcolor}
\usepackage{fancyvrb,fvextra}
\usepackage{etoolbox}

\usepackage{pgf}
\usepackage{collcell}
\usepackage{pgfplots}
\pgfplotsset{compat=1.18}
\newcommand{\dataset}{P3B3\xspace}

\newcommand{\ha}[1]{%
    \cellcolor[rgb]{%
        \fpeval{(#1)/100 > 0.5 ? 0.95 - 0.55*((#1)/100-0.5) : 0.95},
        \fpeval{(#1)/100 > 0.5 ? 0.90 : 0.60 + 0.60*(#1)/100},
        \fpeval{0.55}}#1%
}

\newcommand{\hd}[1]{%
    \cellcolor[rgb]{%
        \fpeval{(#1)/100 > 0.5 ? 0.95 : 0.95 - 0.55*(0.5-(#1)/100)},
        \fpeval{(#1)/100 > 0.5 ? 0.60 + 0.60*(1-(#1)/100) : 0.90},
        \fpeval{0.55}}#1%
}

\newcommand{\hmm}[1]{%
    \cellcolor[rgb]{%
        \fpeval{0.55 - 0.05*(#1)/100},
        \fpeval{0.65 + 0.25*(#1)/100},
        \fpeval{0.95 - 0.40*(#1)/100}}#1%
}

\title{\dataset: A Multi-Turn Conversational Benchmark for Measuring \\ European and Brazilian Portuguese Variety Bias in LLMs}

\author{
 \textbf{Rafael Ferreira\textsuperscript{1,2}},
 \textbf{Inês Vieira\textsuperscript{1}},
 \textbf{Inês Calvo\textsuperscript{1}},
 \textbf{James Furtado\textsuperscript{1,2}},
 \textbf{Iago Paulo\textsuperscript{1,2}},
\\
 \textbf{Diogo Tavares\textsuperscript{1,2}},
 \textbf{Diogo Glória-Silva\textsuperscript{1,2}},
 \textbf{David Semedo\textsuperscript{1,2}},
 \textbf{João Magalhães\textsuperscript{1,2}} \\
 \textsuperscript{1}NOVA University of Lisbon, Portugal, \textsuperscript{2}NOVA LINCS \\
{\small \texttt{\{rah.ferreira, im.paulo, jh.furtado, dc.tavares, dmgc.silva\}@campus.fct.unl.pt}} \\
 {\small \texttt{\{im.vieira, i.calvo, df.semedo, jmag\}@fct.unl.pt}}
}

\begin{document}
\maketitle
\begin{abstract}
As Large Language Models (LLMs) become embedded in everyday communication, capturing regional linguistic variation is essential for reliable and equitable language use. 
In Portuguese, European (pt-PT) and Brazilian (pt-BR) varieties remain unevenly represented, with pt-BR dominating in data quantity, while LLM preference for Portuguese variants remains underexplored.
To address this gap, we introduce \dataset, an expert-curated language variety agnostic benchmark of conversational prompts, along with an evaluation framework for measuring variety bias and controllability.
Experiments on several models show that most LLMs exhibit a strong bias toward pt-BR, with variation in controllability across models. These results highlight the need for more balanced multilingual representation across language varieties.\footnote{\url{https://github.com/Amalia-LLM/p3b3-benchmark}}
\end{abstract}



\section{Introduction}
\label{sec_intro}

LLMs are increasingly used in everyday communication, supporting applications such as customer service, creative writing, and educational tools. As these systems become more widely deployed, their ability to capture regional linguistic variation is critical for reliable and equitable language use, as well as for linguistic and cultural diversity preservation~\cite{vardial_workshop,culturalbench}.

Portuguese is a pluricentric language with over 250 million speakers worldwide. While it shares a common core, the divergence between European Portuguese (pt-PT) and Brazilian Portuguese (pt-BR) is substantial, spanning lexical choices (e.g., \textit{autocarro} vs. \textit{ônibus}), forms of address (e.g., \textit{tu} vs. \textit{você}), and syntax (e.g., clitic placement and gerund usage)~\cite{rio2022portugues,across_atlantic}. Despite these differences, most training and evaluation frameworks treat Portuguese as a single, uniform language~\cite{grattafiori2024llama,qwen3technicalreport,ministral-3}. Given the strong prevalence of Brazilian data in large-scale datasets~\cite{fineweb,CommonCrawl}, this often leads to a silent bias, where models default to pt-BR even when the context suggests pt-PT~\cite{amalia,frmt,iberbench}. This behavior can degrade user experience and contribute to linguistic uneven leveling, where regional variation is underrepresented in LLMs.

\begin{figure}[tb]
    \centering
    \includegraphics[width=0.89\linewidth]{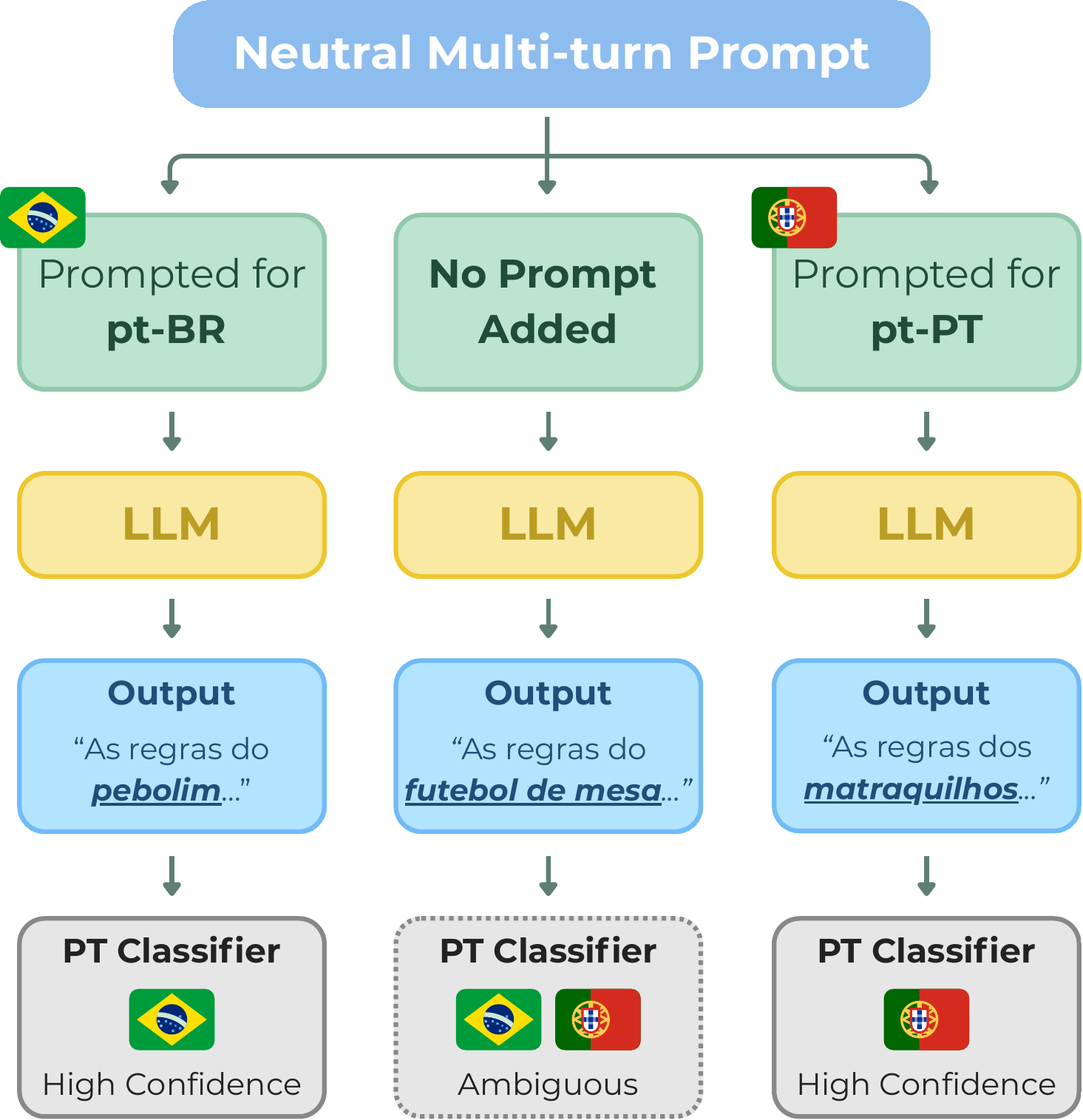}
    \caption{\dataset example and evaluation methodology. The prompts elicit different Portuguese variants, illustrated by the lexical variations of ``\textit{table football}'' in the outputs.}
    \label{fig_p3b3_intro}
\end{figure}

This imbalance raises two key questions: (i) to what extent do current LLMs exhibit a preference for one Portuguese variant, and (ii) can models be steered toward a target variant.

Facing these questions, our contributions are threefold. First, we introduce the pt-PT/pt-BR Bias Benchmark (\dataset), a benchmark of 74 expert-curated, variety-agnostic multi-turn dialogues designed to elicit Portuguese variety-specific linguistic markers in model responses. Second, we propose a human validated evaluation framework combining classifiers~\cite{liad_pt_classifier} and LLM-as-Judge methods~\cite{llm_as_judge_survey} to assess variant alignment and controllability. Finally, we conduct an extensive evaluation of diverse LLMs, revealing a bias toward pt-BR and variable success in steering generation toward pt-PT.

Together, the contributions summarized in Figure~\ref{fig_p3b3_intro} enable the systematic analysis of pt-PT and pt-BR language varieties in LLMs and support the development of more equitable multilingual systems.

\section{Related Work}
\label{sec_related_work}

\paragraph{LLM Performance across Linguistic Varieties.}
A growing body of work shows that LLMs exhibit systematic disparities across linguistic varieties. \citet{dialecticalbiasesinllms} and \citet{transenvframeworkevaluatinglinguistic} show that model performance degrades on non-standard linguistic varieties across a wide range of English dialects and evaluation settings. Extending this line of work, \citet{dialecticpreferencebias} further find a dialectal preference bias, where models assign more negative sentiment to African American English than to equivalent Standard American English inputs.

\paragraph{Portuguese Variant Classification.}
Given the differences across Portuguese varieties~\cite{rio2022portugues}, prior work has studied variant identification using classification-based methods~\cite{liad_pt_classifier,across_atlantic}. 
Furthermore, \citet{ptbrdialectalprofiling} show that LLMs yield inconsistent and unreliable predictions in Brazilian Portuguese dialect profiling. 

\paragraph{Multilingual and Portuguese-Specific Models.}
Most multilingual LLMs treat Portuguese as a monolithic language, without explicit modeling of variations~\cite{qwen3technicalreport,ministral-3,grattafiori2024llama}. 
In response, recent work introduces variant-specific LLMs, including Sabiá~\cite{sabia_3,sabia_4} and Bode~\cite{garcia2024bode} for pt-BR, and GlórIA~\cite{gloria} and AMALIA~\cite{amalia} for pt-PT, aiming to improve cultural and variety alignment.

\vspace{2mm}
\noindent
While prior work examines variety bias in English or frames Portuguese variation as a classification task, we study generation-time bias across Portuguese variants via the expert-designed \dataset, assessing inherent model preferences and their controllability.

\begin{figure}
    \centering
    \includegraphics[width=0.99\linewidth]{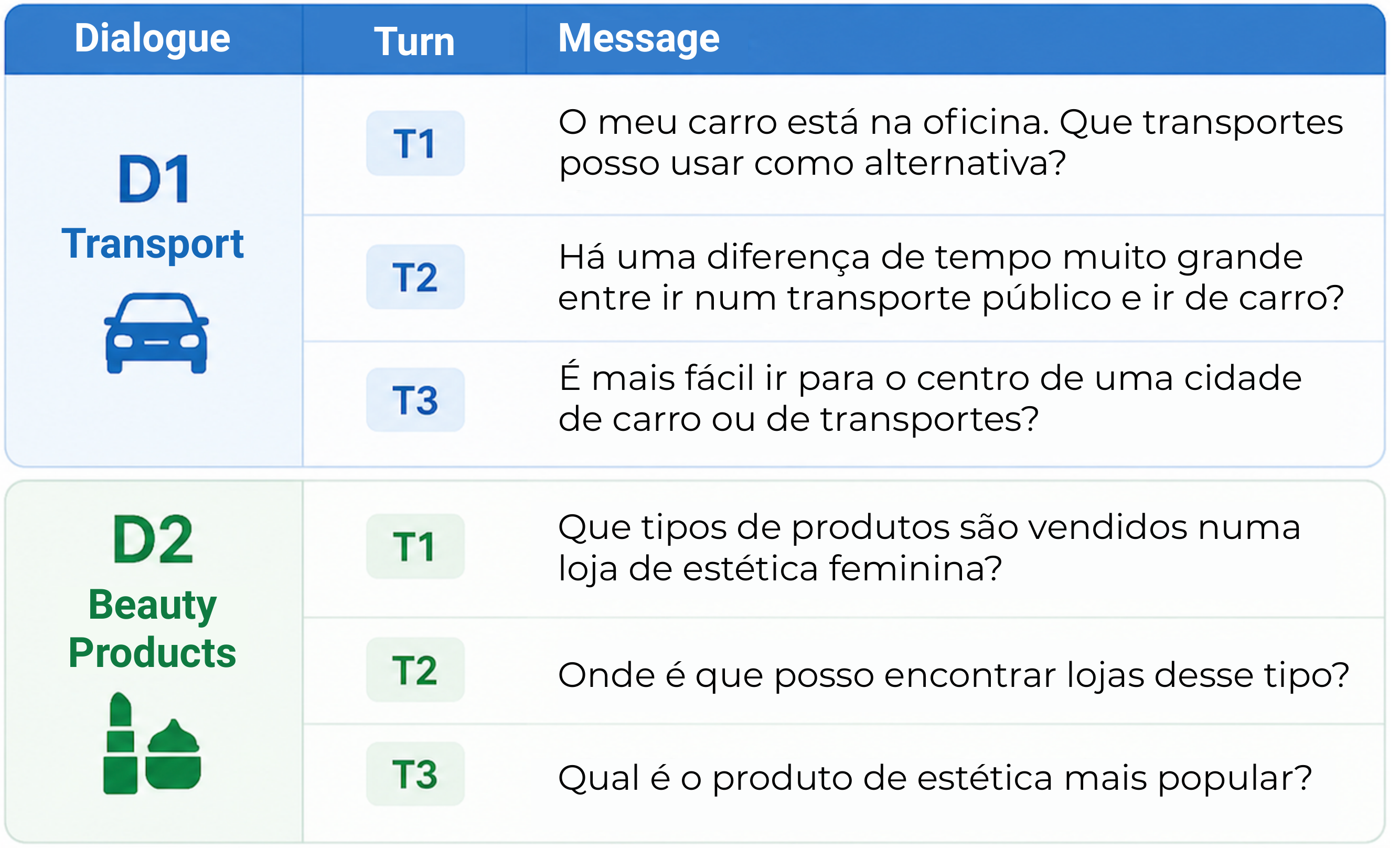}
    \caption{Examples of multi-turn variety-agnostic prompts from \dataset.
    Translation in Figure~\ref{fig_examples_translated}.
    }
    \label{fig_examples}
\end{figure}

\section{The \dataset Benchmark}
\label{sec_p3b3}

The novelty of \dataset lies in systematically evaluating whether LLMs exhibit preferences toward Portuguese variants and the controllability of such behavior. Specifically, \dataset is designed around three guidelines: (i) \textbf{Variety-agnostic}, with no explicit references to variants in the input to prevent priming effects; (ii) \textbf{Variety Sensitivity}, where prompts elicit natural lexical and grammatical differences between Portuguese varieties; and (iii) \textbf{Conversational}, with multi-turn examples that accumulate dialogue history.

These guidelines ensure that variation emerges from generation behavior rather than explicit cues or isolated inputs.

\subsection{Expert-Based Dialogue Creation}
\label{sub_example_creation}

Following MT-Bench~\cite{mt_bench}, \dataset consists of 74 multi-turn dialogues (203 turns in total), each with 2 to 6 turns. As in MT-Bench, later user inputs build on prior dialogue context without requiring specific responses in earlier turns, enabling isolation of generative behavior while maintaining conversational coherence.
This design enables cumulative context effects that may amplify biases not observable in single-turn prompts.

To ensure quality and validity, \dataset was manually constructed by two language experts holding Master’s degrees in linguistics-related fields and with expertise in Portuguese variants. Having this expert knowledge was essential to pinpoint subtle but systematic variation cues, particularly in vocabulary, orthography, and forms of address~\cite{rio2022portugues}, while remaining variant-agnostic.

The benchmark focuses on diverse everyday domains (e.g., transportation, shopping, household products), which exhibit high variability across Portuguese varieties.
These characteristics make \dataset suitable for probing generation-time preferences rather than simple surface-level markers.
Additional benchmark statistics are provided in Appendix~\ref{app_dataset_stats}.

Figure~\ref{fig_examples} shows representative examples from the dataset, which are variety-agnostic and elicit both lexical and syntactic variation in pt-PT and pt-BR. For instance, in Dialogue-1, Turn-1, the prompt elicits lexical variation such as \textit{ônibus} (pt-BR) vs. \textit{autocarro} (pt-PT), as well as grammatical differences in the response.

\section{\dataset Experimental Setup}
\label{sec_evaluation_setup}

\subsection{Priming for Variant Bias and Controllability}
\label{sub_pt_pt_generation}

We evaluate inherent bias and controllability using \dataset under three settings: (i) \textbf{No-Prompt}, where no explicit variety specification is provided, measuring implicit bias; (ii) \textbf{pt-BR Prompt}; and (iii) \textbf{pt-PT Prompt}, where in the first dialogue turn the model is explicitly instructed to respond in the corresponding variant.

Following MT-Bench~\cite{mt_bench}, models generate a response at each turn, with the full dialogue history provided as context.

\subsection{Automatic Evaluation Methods}
\label{sub_pt_br_metrics}

We employ two evaluation methods.

\paragraph{Encoder Classifiers.}

Model outputs are scored using pt-PT/pt-BR binary classifiers. We use \textbf{PeroVaz (PV)}~\cite{Bastos2024PeroVaz} and \textbf{PtBrVId (VId)}~\cite{liad_pt_classifier}, both fine-tuned BERT-based models~\cite{bert_original}.

The encoder outputs a probability score with lower values indicating pt-BR and higher values indicating pt-PT.
Due to input length constraints, we apply a sliding window over each answer and average the scores. 
While efficient, these models may miss subtle or context-dependent variation, motivating the use of LLM-based evaluation.

\paragraph{LLM-as-Judge.}

We use Gemini-3-Flash~\cite{gemini3} to assign a score from 0 (pt-BR) to 10 (pt-PT), along with a chain-of-thought trace~\cite{cot_paper}. We use Gemini-3 due to its reliable instruction-following ability and strong understanding of Portuguese~\cite{globalpiqa}\footnote{In Appendix~\ref{app_gemma_judge}, we show that Gemma-4~\cite{gemma4} can be used as a strong open-source alternative.}.

To improve reliability, we vary three factors: prompt language (\textit{EN} or \textit{PT}), evaluation scope (Single-Turn, \textit{ST}, or full dialogue history, \textit{D}), and prompt detail (Simple, \textit{S}, or Category-based, \textit{C}). Category prompts include 11 expert-defined linguistic categories added to the prompt as examples, such as pronoun placement, preposition contractions, vocabulary, and spelling.

\begin{table}[tb]
\centering
\footnotesize
\resizebox{\linewidth}{!}{%
\begin{tabular}{@{}lccc@{}}
\toprule
\textbf{Method} & \textbf{\begin{tabular}[c]{@{}c@{}}Weighted $\kappa$\\ (Quadratic)\end{tabular}} & \textbf{Pearson} & \textbf{MAE $\downarrow$} \\ \midrule
Humans                          & 0.69          & 0.75          & 2.07          \\ \midrule
PeroVaz (PV)                    & 0.42          & 0.50          & 2.43          \\
PtVId (VId)                     & 0.57          & 0.60          & 2.66          \\ \midrule
Gemini-3 Flash  - EN-S-ST     & 0.79          & 0.81          & 1.79          \\
Gemini-3 Flash - PT-S-ST     & 0.79          & 0.82          & 1.72          \\
Gemini-3 Flash - PT-S-H   & 0.80          & \textbf{0.83}          & 1.72          \\
Gemini-3 Flash - PT-C-ST   & \textbf{0.81} & \textbf{0.83} & \textbf{1.58} \\
Gemini-3 Flash - PT-C-D & \textbf{0.81} & 0.82          & 1.60          \\ 
\bottomrule
\end{tabular}%
}
\caption{Agreement between human annotations and automatic metrics. 
EN/PT = language; S = Simple; C = Category; ST = Single-Turn; D = Dialogue history.}
\label{tab_judge_metrics}
\end{table}

\subsubsection{Validating Automatic Metrics}
We validate the metrics using 200 responses annotated by two linguistic experts on a 0 (pt-BR) to 10 (pt-PT) scale, with -1 for invalid outputs. Samples are randomly drawn equally from all settings (No-Prompt, pt-BR, pt-PT) across 12 models.
Considering valid responses (88.5\%), Table~\ref{tab_judge_metrics} reports agreement with human judgments.

Human annotators show strong agreement, indicating task reliability.
Classifiers achieve moderate alignment with humans, with VId performing best. The LLM-as-a-Judge exhibits strong alignment, performing best under the PT-C-ST setup, which is used in subsequent experiments.

Overall, classifiers provide a low-cost complementary signal, while LLM-based evaluation yields more aligned and interpretable results.

\subsection{Models and Implementation Details}
\label{sub_models_and_details}

We evaluate a range of multilingual instruction-tuned LLMs of varying sizes (list in Appendix~\ref{app_model_list}). Models are selected based on multilingual and Portuguese performance, public availability, and architectural diversity.
All models are evaluated in BF16 using greedy decoding.
Additionally, we include closed-source models: Sabiá-4~\cite{sabia_4}, specialized for Brazilian Portuguese, and Gemini-3-Flash~\cite{gemini3} as a frontier model.

\section{Results and Discussion}
\label{sec_results}

\begin{table}[tbp]
\centering
\footnotesize
\resizebox{\linewidth}{!}{%
\begin{tabular}{@{}l|ccc|ccc|ccc@{}}
\toprule
                                             & \multicolumn{3}{c|}{\textbf{No-Prompt}} & \multicolumn{3}{c|}{\textbf{pt-BR ($\downarrow$)}}    & \multicolumn{3}{c}{\textbf{pt-PT ($\uparrow$)}}     \\ \midrule
\textbf{Model}                                        & \textbf{PV}  & \textbf{VId} & \textbf{LLM}  & \textbf{PV} & \textbf{VId} & \textbf{LLM}  & \textbf{PV} & \textbf{VId} & \textbf{LLM}  \\ \midrule
\textbf{\textit{Fully Open}}                                      &        &          &             &      &        &          &             &      &        \\
Salamandra-7B                & \hmm{47.9}      & \hmm{32.8}        & \hmm{34.5} &  \hd{44.9}     & \hd{22.1}        & \hd{20.7}    & \ha{52.0}     & \ha{43.7}        & \ha{52.8}     \\
OLMo-3-7B                  & \hmm{38.6}      & \hmm{37.0}        & \hmm{17.7}     & \hd{37.5}     & \hd{29.4}        & \hd{15.8}   & \ha{45.8}     & \ha{41.0}        & \ha{23.3}    \\
OLMo-3.1-32B                & \hmm{38.9}      & \hmm{22.2}        & \hmm{16.6}    & \hd{32.9}     & \hd{8.0}         & \hd{10.5}   & \ha{58.9}     & \ha{51.9}        & \ha{42.8}    \\
EuroLLM-22B      & \hmm{38.1}      & \hmm{25.3}        & \hmm{22.5}    & \hd{35.6}     & \hd{10.7}        & \hd{9.6}     & \ha{70.3}     & \ha{87.7}        & \ha{88.5}     \\
Apertus-8B           & \hmm{43.2}      & \hmm{26.8}        & \hmm{20.3}     & \hd{41.9}     & \hd{18.9}        & \hd{14.7} &  \ha{59.6}     & \ha{64.1}        & \ha{61.0}    \\
Apertus-70B           & \hmm{41.0}      & \hmm{25.7}        & \hmm{16.1}   & \hd{36.5}     & \hd{13.6}        & \hd{8.6}  &  \ha{68.8}     & \ha{85.7}        & \ha{86.3}     \\
AMALIA-9B               & \hmm{66.9}      & \hmm{85.6}        & \hmm{91.2}    & \hd{45.9}     & \hd{33.5}        & \hd{34.8}    & \ha{71.3}     & \ha{91.8}        & \ha{95.9}    \\ \midrule
\textbf{\textit{Open Weight}}                  &           &             &      &        &          &             &      &        &                                \\
Llama-3.1-8B            & \hmm{41.0}      & \hmm{24.1}        & \hmm{12.6}    & \hd{38.6}     & \hd{17.0}        & \hd{10.6}     & \ha{57.1}     & \ha{69.3}        & \ha{47.7}    \\
Llama-3.3-70B            & \hmm{40.9}      & \hmm{25.6}        & \hmm{15.3}     & \hd{34.6}     & \hd{13.4}        & \hd{6.7}    & \ha{61.6}     & \ha{81.5}        & \ha{67.8}    \\
Gervásio-8B-ptpt & \hmm{39.7}      & \hmm{25.7}        & \hmm{12.7}    & \hd{38.6}     & \hd{17.8}        & \hd{11.1}    & \ha{55.3}     & \ha{65.5}        & \ha{46.6}    \\
Gervásio-70B-ptpt & \hmm{47.4}      & \hmm{32.4}        & \hmm{22.8}     & \hd{36.8}     & \hd{16.7}        & \hd{10.8}     & \ha{64.6}     & \ha{84.8}        & \ha{75.2}    \\
Ministral-3-14B      & \hmm{43.6}      & \hmm{23.3}        & \hmm{17.8}    & \hd{32.5}     & \hd{6.9}         & \hd{4.2}     & \ha{74.8}     & \ha{95.2}        & \ha{90.7}     \\
Qwen3-8B                                & \hmm{42.0}      & \hmm{15.7}        & \hmm{10.4}    & \hd{36.4}     & \hd{9.1}         & \hd{7.9}     & \ha{48.1}     & \ha{41.0}        & \ha{32.7}    \\
Qwen3.5-9B                              & \hmm{42.9}      & \hmm{42.0}        & \hmm{35.1}     & \hd{28.7}     & \hd{12.8}        & \hd{12.2}     & \ha{76.0}     & \ha{93.6}        & \ha{86.1}    \\
Qwen3.5-27B                             & \hmm{47.5}      & \hmm{47.4}        & \hmm{44.5}     & \hd{28.5}     & \hd{11.6}        & \hd{11.9}     & \ha{78.4}     & \ha{98.1}        & \ha{93.3}   \\
Gemma-3-12b                        & \hmm{38.9}      & \hmm{26.5}        & \hmm{23.4}     & \hd{30.3}     & \hd{6.7}         & \hd{4.9}      & \ha{72.2}     & \ha{95.0}        & \ha{92.9}   \\
Gemma-4-E4B                        & \hmm{40.4}      & \hmm{29.3}        & \hmm{21.9}     & \hd{30.1}     & \hd{9.3}         & \hd{4.2}      & \ha{70.1}     & \ha{96.4}        & \ha{94.4}    \\
Gemma-4-31B                        & \hmm{49.2}      & \hmm{52.3}        & \hmm{49.2}     & \hd{28.4}     & \hd{8.3}         & \hd{7.3}    & \ha{69.7}     & \ha{95.6}        & \ha{98.0}     \\ \midrule
\textbf{\textit{Close Source} }                &           &             &      &        &          &             &      &        &                               \\
Sabiá-4                         & \hmm{34.5}      & \hmm{7.0}         & \hmm{8.1}     & \hd{32.5}     & \hd{4.5}         & \hd{4.6}    & \ha{67.0}     & \ha{90.4}        & \ha{87.2}     \\ 
Gemini-3-Flash                         & \hmm{55.5}      & \hmm{63.1}         & \hmm{63.7}     & \hd{23.6}     & \hd{5.8}         & \hd{4.1}      & \ha{79.7}     & \ha{99.3}        & \ha{99.8}     \\ \bottomrule
\end{tabular}%
}
\caption{Results across models and prompting settings in \dataset. All scores are scaled to 0 (pt-BR) -- 100 (pt-PT) measuring variant alignment.}
\label{tab_results}
\end{table}

\subsection{Bias and Controllability Analysis}
\label{sub_bias_analysis}

Table~\ref{tab_results} reports model results across three settings. We note that we measure \textit{variant preference} in generation, rather than response quality; therefore, lower or higher scores indicate a tendency toward pt-BR or pt-PT, respectively, rather than differences in output quality.

\paragraph{No-Prompt.}
This setting measures inherent model bias, showing that most models exhibit a consistent preference for pt-BR, likely reflecting training data imbalance.
AMALIA, which is specifically pre-trained and fine-tuned for pt-PT, is the only model consistently biased toward pt-PT (91.2). In contrast, Gervásio, a LLaMA model fine-tuned for pt-PT, still exhibits a pt-BR bias, suggesting that pre-training data and scale play a critical role in capturing variety-specific behavior. Interestingly, larger and more recent models such as Qwen-3.5 27B, Gemma-4-31B, and Gemini-3-Flash show more balanced behavior across varieties (44.5 to 63.7).

\paragraph{pt-BR.}
When prompted for pt-BR, most models maintain or increase their use of the pt-BR variety, indicating that models are able to follow explicit variety instructions, particularly at larger model sizes.
In contrast, AMALIA is an outlier showing a higher tendency toward pt-PT even in this setting, consistent with its pt-PT specialized training setup.

\paragraph{pt-PT.}

Results for pt-PT are more variable. Strong alignment is observed from EuroLLM, Apertus-70B, and AMALIA ($\geq$ 86), while LLaMA-based models struggle to produce consistent pt-PT outputs. All Gemma models perform well in this setting ($\geq$ 92). Notably, Qwen-3.5 shows a substantial improvement over Qwen-3 (32.7 to 86.1), suggesting that changes in training significantly improved pt-PT handling. Despite being pt-BR-oriented, Sabiá-4 generalizes well to pt-PT (87.2).

\paragraph{Classifiers vs. LLM Scores.}

Classifier-based and LLM-based evaluations yield broadly consistent rankings. Overall, classifiers provide a low-cost complementary signal, while LLM-based evaluation serves as the primary analysis due to higher accuracy and interpretability.

\begin{figure}[t]
    \centering
    \includegraphics[width=0.99\linewidth]{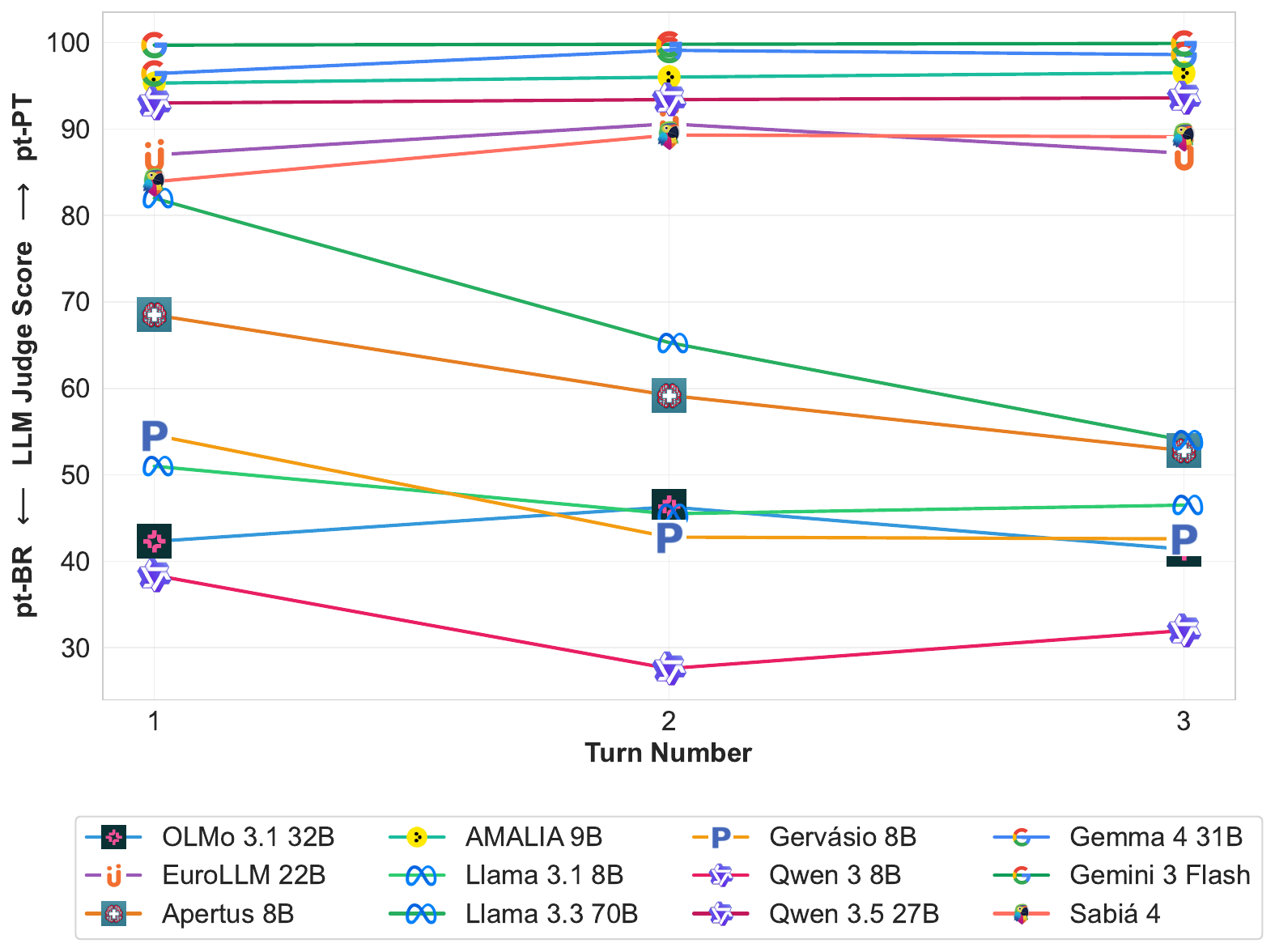}
    \caption{Alignment across turns for pt-PT prompt.}
    \label{fig_turn_progression_pt_pt}
\end{figure}

\subsection{Turn-level Analysis}
\label{sub_turn_level_analysis}

Figure~\ref{fig_turn_progression_pt_pt} shows LLM scores across the first three dialogue turns (later turns are excluded due to sample size).
We focus on pt-PT as it has the most variance.
Except for Llama, we observe that models with strong initial pt-PT alignment generally maintain consistent behavior across turns. 
Models with lower initial tendency to pt-PT tend to drift toward pt-BR over time. This suggests that a single initial instruction is often insufficient to sustain language variety adherence across multiple turns, indicating that controllability degrades in longer contexts.
For the No-Prompt and pt-BR settings, behavior remains stable across turns (Appendix~\ref{app_eval_by_turn}).

\subsection{Human Analysis}

Linguistic experts evaluated a random subset of various model outputs in the more challenging pt-PT setting.
As observed in the automatic metrics, weaker models tend to drift toward pt-BR, exhibiting proclisis, gerund constructions, pronominal usage (notably \textit{você}), and pt-BR lexical choices (e.g. \textit{geladeira} and \textit{ônibus}). They also show inconsistencies in orthography and accentuation, often producing hybrid outputs that mix pt-BR grammar with partial pt-PT vocabulary~\cite{rio2022portugues}.
In contrast, stronger models largely follow pt-PT conventions, indicating better syntactic and morphological control.
We provide examples of model and judge outputs in Appendix~\ref{app_example_outputs}.

\section{Conclusion}
\label{sec_conclusion}

We introduced \dataset, a multi-turn benchmark for evaluating bias and controllability in LLM generation across Portuguese variants.
Most models exhibit a strong bias toward pt-BR, and while LLMs can be steered toward different varieties, larger and newer models show greater controllability. However, pt-PT performance remains less consistent across models, highlighting the need for better representation of linguistic variation in training and evaluation.

\section*{Limitations}
\label{sec_limitations}

\dataset provides a systematic evaluation of LLM adherence to Portuguese varieties in conversational domains. The current setup does not include technical or specialized domains, focusing on everyday topics. Moreover, while \dataset supports analysis of pt-BR and pt-PT, extending it to other Portuguese varieties spoken in countries such as Angola, Mozambique, and Cape Verde is a promising direction for future work.

\section*{Acknowledgments}
This work was supported by the AMALIA project under Measure RE-C05-i08 of the Portuguese national Programa de Recuperação e Resiliência. We also acknowledge the support of Fundação para a Ciência e Tecnologia (FCT) and the NOVA LINCS project (UID/04516/2025). Finally, we thank the Barcelona Supercomputing Center (BSC) for providing the computational resources that made this work possible.

\bibliography{references}

\appendix

\section{\dataset Benchmark Statistics}
\label{app_dataset_stats}
\dataset was constructed through a manual 
process conducted by two language experts with expertise in Portuguese and comprises of 74 conversations and 203 turns.

\paragraph{Structural Distribution.}
The dataset targets short-form multi-turn interactions with dialogue turns distributed as follows: 31 dialogues have 2 turns, 35 have 3, 5 have 4, and 3 have 5--6 turns.

\paragraph{Lexical and Sequence Diversity.}
The dataset shows strong lexical diversity. Per-conversation TTR is 0.73, with Distinct-2 and Distinct-3 scores of 0.71 and 0.88, indicating high phrasal variation. Lower Distinct-1 (0.26) and global TTR (0.25) reflect expected function word frequency. Entropy increases from 7.7 (1-gram) to 10.9 (3-gram), confirming linguistic variability via different conversational domains.

\paragraph{Length Statistics.}
Turns average 14.1 words (SD=6.0), with word counts averaging 38.7 per dialogue and ranging from 14 to 74 words, consistent with the benchmark's focused design.

\section{Detailed Model List}
\label{app_model_list}
Table~\ref{tab_model_paths} provides a comprehensive list of the models evaluated in this work, along with their respective model identifiers.

\begin{table}[htb]
    \centering
    \footnotesize
    \resizebox{\linewidth}{!}{%
    \begin{tabular}{ll}
        \toprule
        \textbf{Model Provider \& Name} & \textbf{Model Identifier} \\
        \midrule
        \multicolumn{2}{l}{\textbf{BSC-LT}} \\
        Salamandra 7B~\cite{gonzalez2025salamandra} & \texttt{salamandra-7b-instruct} \\
        \midrule
        \multicolumn{2}{l}{\textbf{allenai}} \\
        OLMo 3 7B~\cite{olmo3} & \texttt{Olmo-3-7B-Instruct} \\
        OLMo 3.1 32B~\cite{olmo3} & \texttt{Olmo-3.1-32B-Instruct} \\
        \midrule
        \multicolumn{2}{l}{\textbf{utter-project}} \\
        EuroLLM 22B~\cite{eurollm} & \texttt{EuroLLM-22B-Instruct-2512} \\
        \midrule
        \multicolumn{2}{l}{\textbf{swiss-ai}} \\
        Apertus 8B~\cite{apertus} & \texttt{Apertus-8B-Instruct-2509} \\
        Apertus 70B~\cite{apertus} & \texttt{Apertus-70B-Instruct-2509} \\
        \midrule
        \multicolumn{2}{l}{\textbf{AMALIA-LLM}} \\
        AMALIA 9B~\cite{amalia} & \texttt{AMALIA-9B-50-1225-DPO} \\
        \midrule
        \multicolumn{2}{l}{\textbf{meta-llama}} \\
        Llama 3.1 8B Instruct~\cite{grattafiori2024llama} & \texttt{Llama-3.1-8B-Instruct} \\
        Llama 3.3 70B~\cite{grattafiori2024llama} & \texttt{Llama-3.3-70B-Instruct} \\
        \midrule
        \multicolumn{2}{l}{\textbf{PORTULAN}} \\
        Gervásio 8B-pt-pt~\cite{gervasio} & \texttt{gervasio-8b-portuguese-ptpt-decoder} \\
        Gervásio 70B-pt-pt~\cite{gervasio} & \texttt{gervasio-70b-portuguese-ptpt-decoder} \\
        \midrule
        \multicolumn{2}{l}{\textbf{mistralai}} \\
        Ministral 3 14B~\cite{ministral-3} & \texttt{Ministral-3-14B-Instruct-2512} \\
        \midrule
        \multicolumn{2}{l}{\textbf{Qwen}} \\
        Qwen 3 8B~\cite{qwen3technicalreport} & \texttt{Qwen3-8B} \\
        Qwen 3.5 9B & \texttt{Qwen3.5-9B} \\
        Qwen 3.5 27B & \texttt{Qwen3.5-27B} \\
        \midrule
        \multicolumn{2}{l}{\textbf{google}} \\
        Gemma 3 12B~\cite{gemma3} & \texttt{gemma-3-12b-it} \\
        Gemma 4 E4B~\cite{gemma4} & \texttt{gemma-4-E4B-it} \\
        Gemma 4 31B~\cite{gemma4} & \texttt{gemma-4-31B-it} \\
        \midrule
        \multicolumn{2}{l}{\textbf{API-only}} \\
        Gemini 3 Flash Preview~\cite{gemini3} & \texttt{Google API - gemini-3-flash-preview} \\
        Sabiá 4~\cite{sabia_4} & \texttt{Maritaca API - sabia-4} \\
        \bottomrule
    \end{tabular}
    }
    \caption{Model names and identifiers used in this work.}
    \label{tab_model_paths}
\end{table}

\section{Turn-level Analysis}
\label{app_eval_by_turn}

\begin{figure}[h]
    \centering
        \begin{minipage}{0.98\linewidth}
        \centering
        \includegraphics[width=\linewidth]{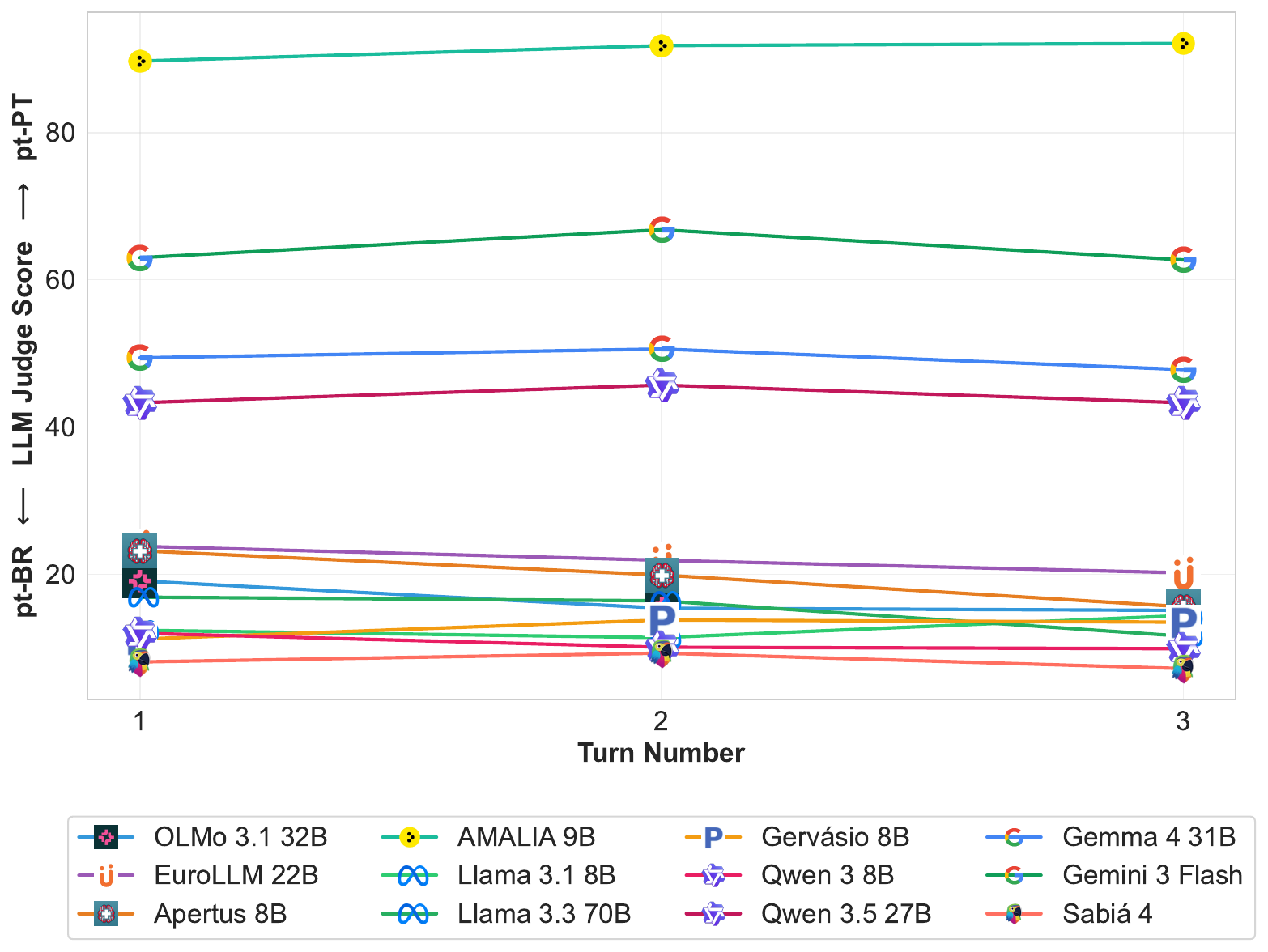}
        \caption{Alignment across turns for No-Prompt.}
        \label{fig_turn_progression_normal}
    \end{minipage}
    \\
    \begin{minipage}{0.98\linewidth}
        \centering
        \includegraphics[width=\linewidth]{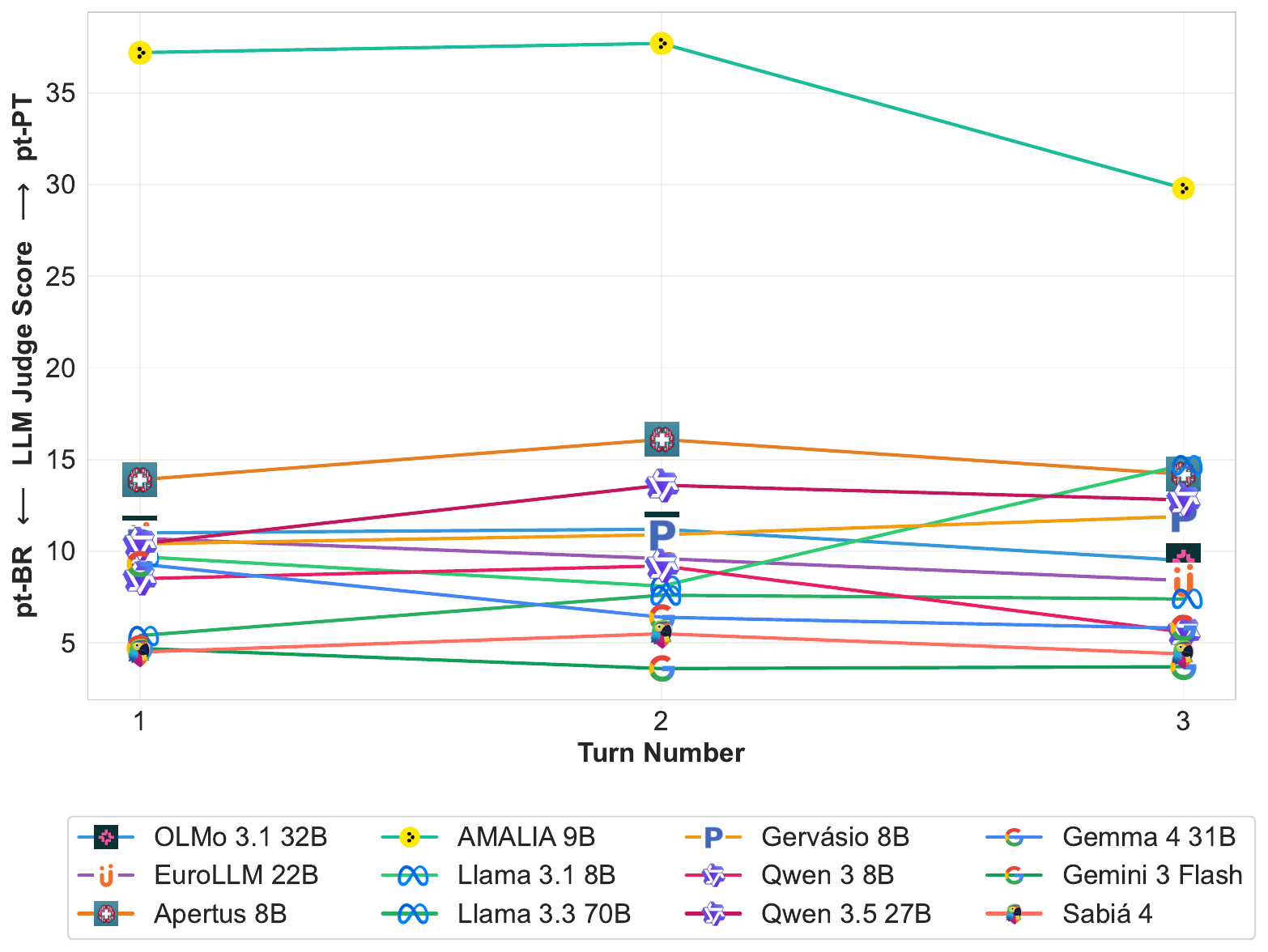}
        \caption{Alignment across turns for pt-BR prompt.}
        \label{fig_turn_progression_pt_br}
    \end{minipage}
\end{figure}

Figures~\ref{fig_turn_progression_normal} and \ref{fig_turn_progression_pt_br} show LLM scores across the first three dialogue turns for the No-Prompt and pt-BR settings, respectively.
In contrast to the pt-PT setting (Figure~\ref{fig_turn_progression_pt_pt}), where alignment progressively shifts toward pt-BR, these configurations exhibit largely stable behavior across turns. This consistency suggests a persistent model bias toward pt-BR, which remains unchanged throughout the interaction, with no tendency to move toward pt-PT.

\begin{figure}[tbp]
    \centering
    \includegraphics[width=0.96\linewidth]{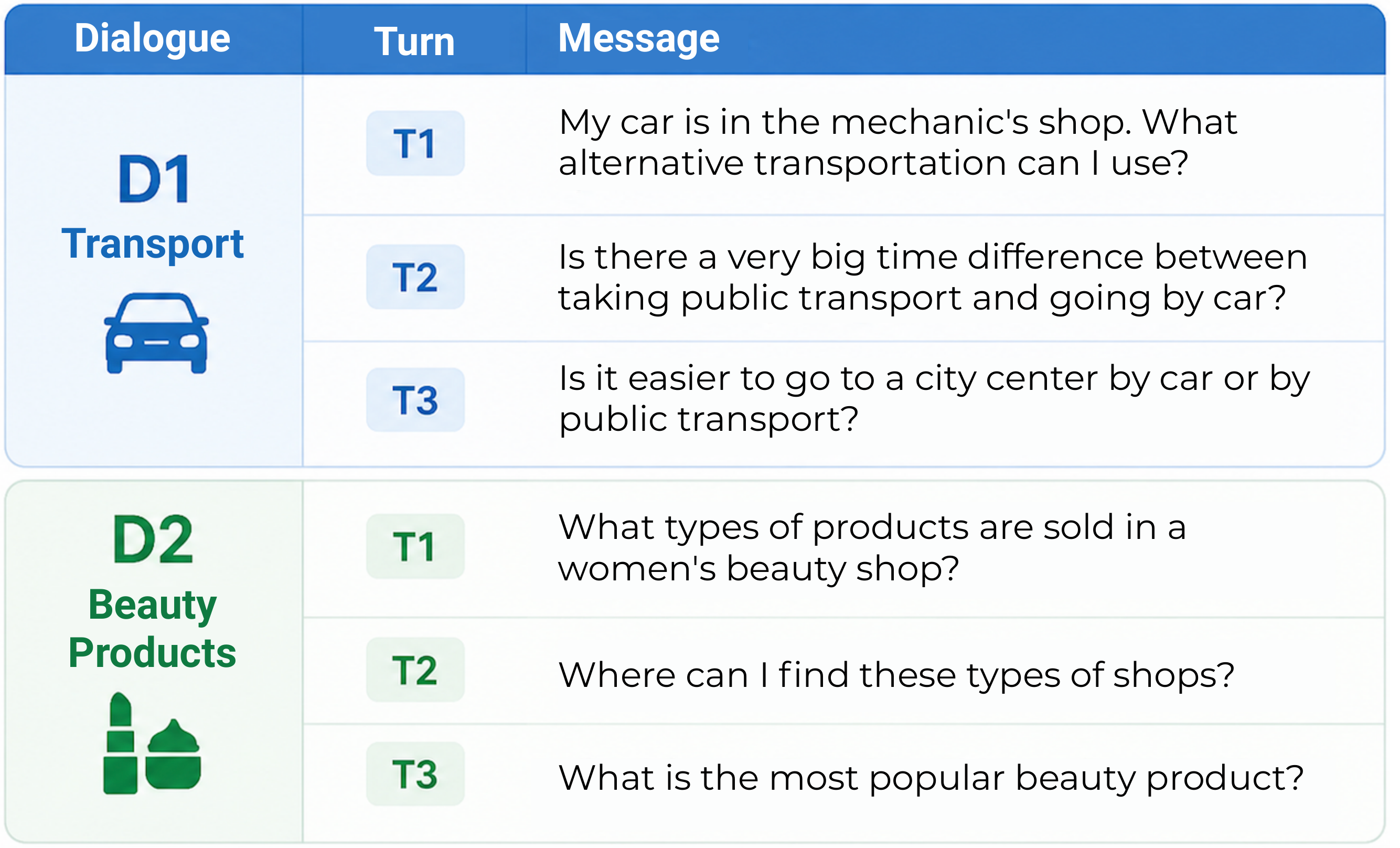}
    \caption{Translated examples from \dataset benchmark. Original Portuguese version in Figure~\ref{fig_examples}.
    }
    \label{fig_examples_translated}
\end{figure}

\section{Portuguese Output Consistency}
We used Gemini-3 Flash to classify whether outputs were predominantly in Portuguese, regardless of variant.
Most models consistently generate outputs in Portuguese (99\%). The only exception is OLMo-3-7B, which still produces Portuguese in 94\% of cases, with occasional outputs in Spanish.

\section{Extending the LLM-as-Judge Setup}
\label{app_gemma_judge}

We additionally include Gemma-4-31B~\cite{gemma4} in our LLM-as-Judge setup in Table~\ref{tab_judge_metrics_gemma}. Gemma-4-31B performs strongly across all settings and metrics. While it trails Gemini-3-Flash, it significantly improves over the classifiers and serves as a strong open-source alternative for this task, with competitive performance.

With respect to the results, Table~\ref{tab_results_llms} compares the evaluations produced by Gemini-3-Flash and Gemma-4-31B across all models and prompting settings in \dataset. We observe a strong agreement between the two judges, both in ranking order and overall score ranges, with only a few discrepancies in specific settings. These findings further validate our results and highlight the potential of using open-weight LLM judges to reliably measure Portuguese variant alignment.

\section{Example Model and Judge Outputs}
\label{app_example_outputs}

\begin{figure*}[tbh]
    \centering
    \includegraphics[width=0.58\linewidth]{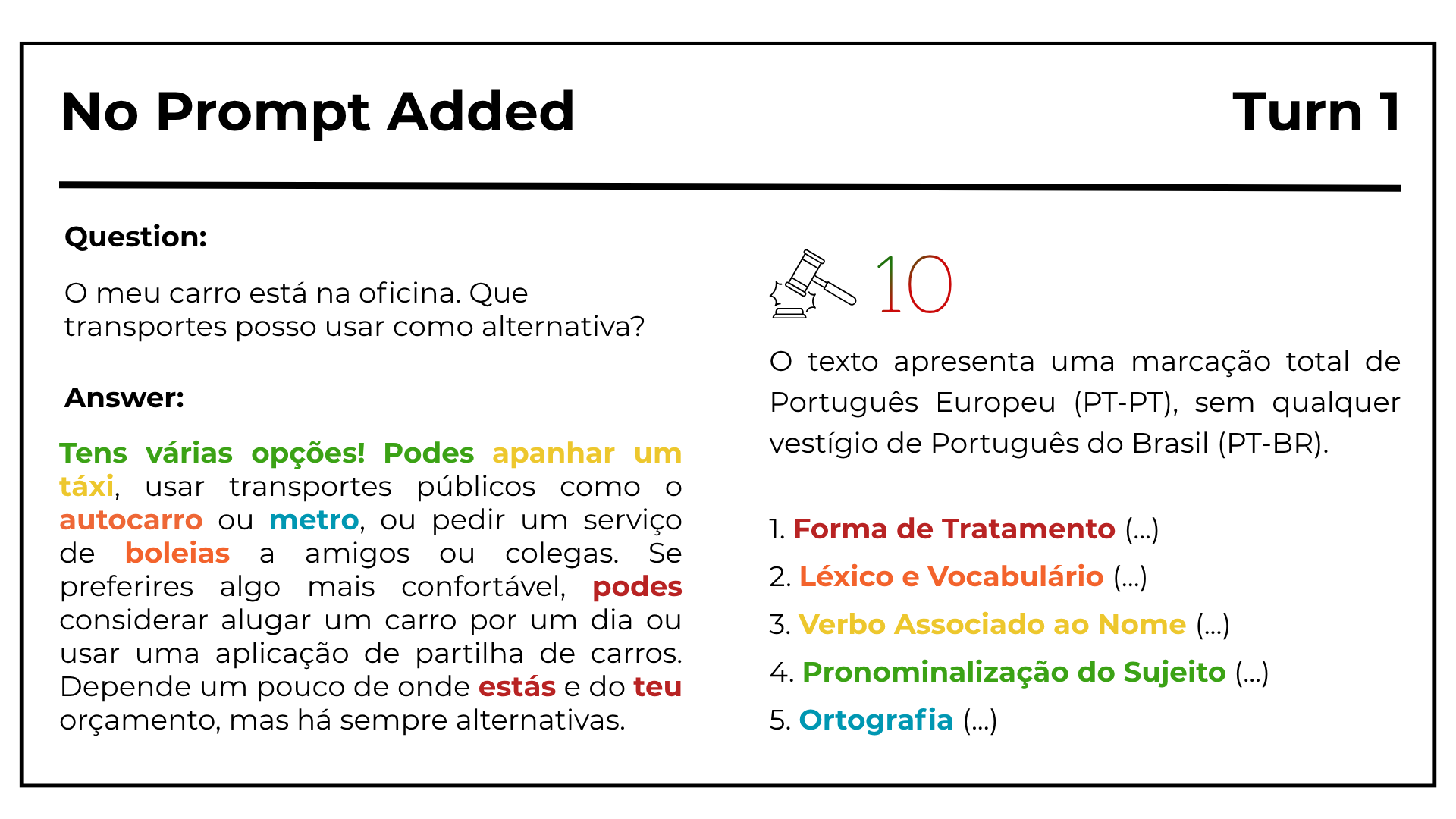}
    \caption{Example of the LLM Judge score and reasoning given to an answer from AMALIA-9B for No-Prompt setting. The colors represent the points in the answer the judge's reasoning is addressing.\\\\\scriptsize\textbf{EN Translation}:
    \textit{\textbf{Question:}
My car is in the shop. What alternative transportation can I use?
\textbf{Answer:}
You have several options! You can catch a taxi, use public transport such as the bus or metro, or ask friends or colleagues for a ride. If you prefer something more comfortable, you can consider renting a car for a day or using a car-sharing app. It depends a bit on where you are and your budget, but there are always alternatives.
\textbf{LLM Judge:}
Score: 10
The text presents a total marking of European Portuguese (PT-PT), without any trace of Brazilian Portuguese (PT-BR).
1. Form of Address (...)
2. Lexicon and Vocabulary (...)
3. Verb Associated with the Noun (...)
4. Pronominalization of the Subject (...)
5. Spelling (...)}}
    \label{fig_outputs_normal}
\end{figure*}

\begin{figure*}
    \centering
    \includegraphics[width=0.58\linewidth]{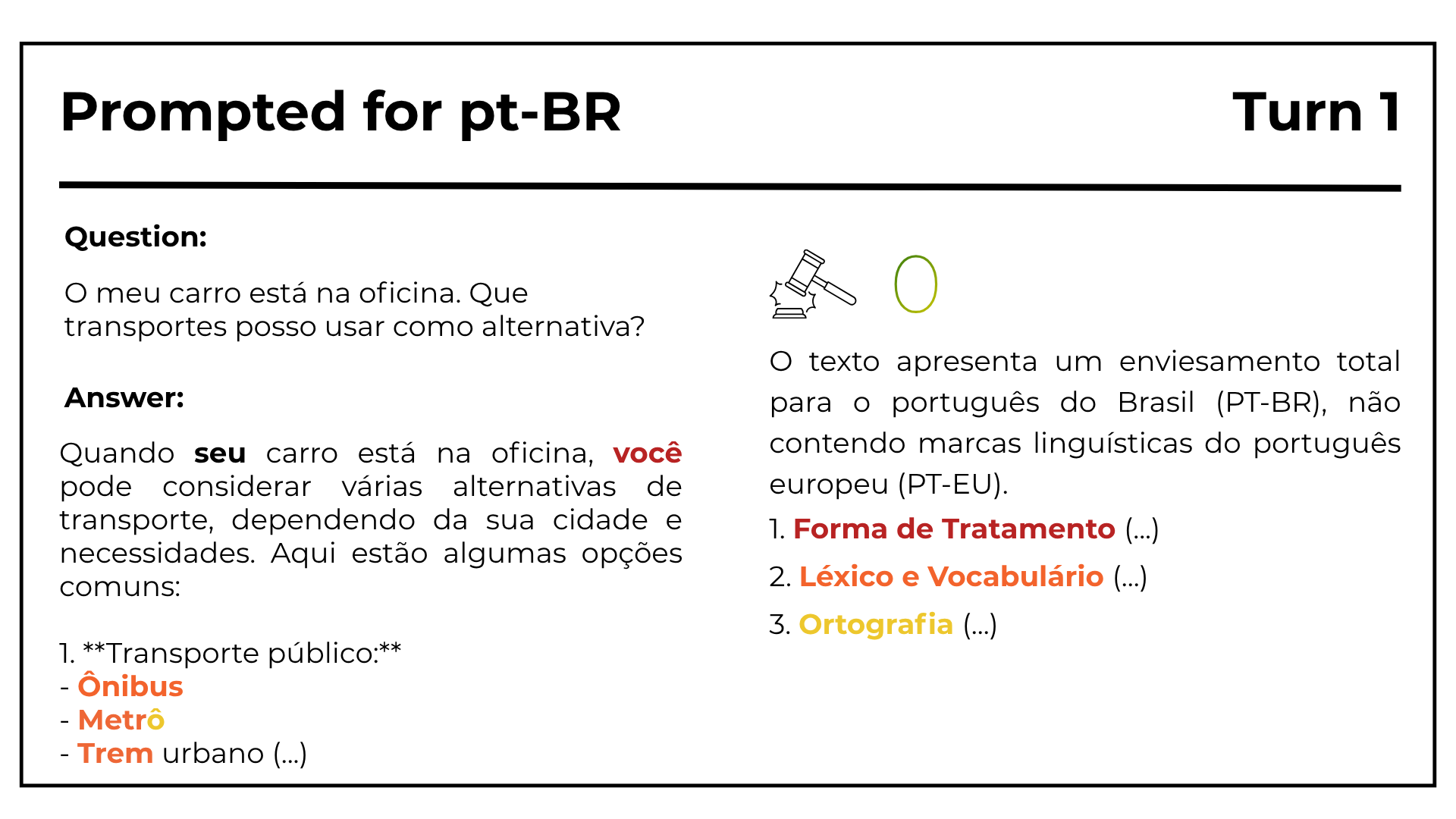}
    \caption{Example of the LLM Judge score and reasoning given to an answer from Sabiá-4 for a pt-BR prompt setting. The colors represent the points in the answer the judge's reasoning is addressing.\\\\\scriptsize\textbf{EN Translation}:
    \textit{\textbf{Question:}
My car is in the repair shop. What transport can I use as an alternative?
\textbf{Answer:}
When your car is in the shop, you can consider various transport alternatives, depending on your city and needs. Here are some common options:
1. Public Transport:
Bus
Subway
Urban train (...)
\textbf{LLM Judge:}
Score: 0
The text shows a total bias toward Brazilian Portuguese (PT-BR), containing no linguistic markers of European Portuguese (PT-EU).
1. Form of Address (...)
2. Lexicon and Vocabulary (...)
3. Spelling (...)
}}
    \label{fig_outputs_pt_br}
\end{figure*}

\begin{figure*}
    \centering
    \includegraphics[width=0.58\linewidth]{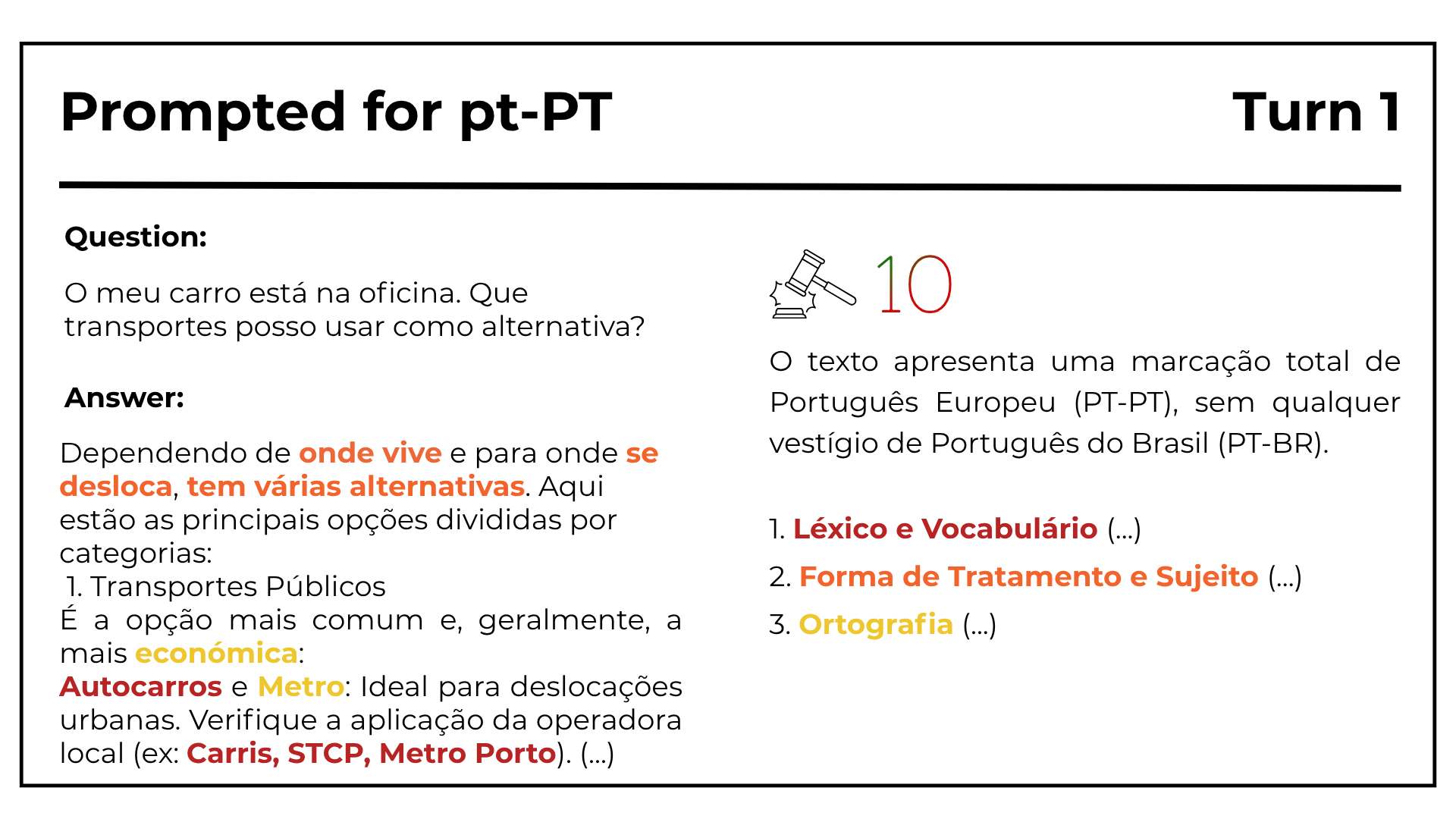}
    \caption{Example of the LLM Judge score and reasoning given to an answer from Gemma-4-31B for a pt-PT prompt setting. The colors represent the points in the answer the judge's reasoning is addressing.\\\\\scriptsize\textbf{EN Translation}:
    \textit{\textbf{Question:}
My car is in the repair shop. What transport can I use as an alternative?
\textbf{Answer:}
Depending on where you live and where you travel to, you have several alternatives. Here are the main options divided by categories:
1. Public Transport
It is the most common and, generally, the most economical option:
Buses and Metro: Ideal for urban travel. Check the app of the local operator (ex: Carris, STCP, Metro Porto). (...)
\textbf{LLM Judge:}
Score: 10
The text presents a total marking of European Portuguese (PT-PT), without any trace of Brazilian Portuguese (PT-BR).
1. Lexicon and Vocabulary (...)
2. Form of Address and Subject (...)
3. Spelling (...)
}}
    \label{fig_outputs_pt_pt}
\end{figure*}

Figures~\ref{fig_outputs_normal}, \ref{fig_outputs_pt_br}, and \ref{fig_outputs_pt_pt} show example model and judge outputs across different settings.

\vfill \pagebreak

\begin{table}[h]
\centering
\footnotesize
\resizebox{\linewidth}{!}{%
\begin{tabular}{@{}lccc@{}}
\toprule
\textbf{Method} & \textbf{\begin{tabular}[c]{@{}c@{}}Weighted $\kappa$\\ (Quadratic)\end{tabular}} & \textbf{Pearson} & \textbf{MAE $\downarrow$} \\ \midrule
Humans                          & 0.69          & 0.75          & 2.07          \\ \midrule
PeroVaz (PV)                    & 0.42          & 0.50          & 2.43          \\
PtVId (VId)                     & 0.57          & 0.60          & 2.66          \\ \midrule
Gemini-3 Flash - EN-S-ST     & 0.79          & 0.81          & 1.79          \\
Gemini-3 Flash - PT-S-ST     & 0.79          & 0.82          & 1.72          \\
Gemini-3 Flash - PT-S-H   & 0.80          & \textbf{0.83}          & 1.72          \\
Gemini-3 Flash - PT-C-ST   & \textbf{0.81} & \textbf{0.83} & \textbf{1.58} \\
Gemini-3 Flash - PT-C-D & \textbf{0.81} & 0.82          & 1.60          \\ 
\midrule
Gemma-4-31B - EN-S-ST     &    0.75      &    0.78       &   1.95        \\
Gemma-4-31B - PT-S-ST     &    0.78      &    0.80       &  1.73          \\
Gemma-4-31B - PT-S-H   &    0.78       &    0.80       &   1.74         \\
Gemma-4-31B - PT-C-ST   & 0.78 & 0.79 & 1.72 \\
Gemma-4-31B - PT-C-D &  0.78 &    0.80       &  1.71         \\ 
\bottomrule
\end{tabular}%
}
\caption{Agreement between human annotations and LLM-based metrics. 
EN/PT = language; S = Simple; C = Category; ST = Single-Turn; D = Dialogue history.}
\label{tab_judge_metrics_gemma}
\end{table}

\begin{table}[h]
\centering
\footnotesize
\resizebox{\linewidth}{!}{%
\begin{tabular}{@{}lcccccc@{}}
\toprule
 &
  \multicolumn{2}{c}{\textbf{No-Prompt}} &
  \multicolumn{2}{c}{\textbf{pt-BR $(\downarrow)$}} &
  \multicolumn{2}{c}{\textbf{pt-PT ($\uparrow$)}} \\ \midrule
\multicolumn{1}{l|}{\textbf{Model}} &
  \textbf{Gemma-4} &
  \multicolumn{1}{c|}{\textbf{Gemini-3}} &
  \textbf{Gemma-4} &
  \multicolumn{1}{c|}{\textbf{Gemini-3}} &
  \textbf{Gemma-4} &
  \textbf{Gemini-3} \\ \midrule
\multicolumn{1}{l|}{\textit{\textbf{Fully Open}}} &
   &
  \multicolumn{1}{c|}{} &
   &
  \multicolumn{1}{c|}{} &
   &
   \\
\multicolumn{1}{l|}{Salamandra-7B} &
  \hmm{35.5} &
  \multicolumn{1}{c|}{\hmm{34.5}} &
  \hd{24.4} &
  \multicolumn{1}{c|}{\hd{20.7}} &
  \ha{54.4} &
  \ha{52.8} \\
\multicolumn{1}{l|}{OLMo-3-7B} &
  \hmm{23.1} &
  \multicolumn{1}{c|}{\hmm{17.7}} &
  \hd{18.1} &
  \multicolumn{1}{c|}{\hd{15.8}} &
  \ha{25.6} &
  \ha{23.3} \\
\multicolumn{1}{l|}{OLMo-3.1-32B} &
  \hmm{17.0} &
  \multicolumn{1}{c|}{\hmm{16.6}} &
  \hd{13.3} &
  \multicolumn{1}{c|}{\hd{10.5}} &
  \ha{47.7} &
  \ha{42.8} \\
\multicolumn{1}{l|}{EuroLLM-22B} &
  \hmm{23.4} &
  \multicolumn{1}{c|}{\hmm{22.5}} &
  \hd{11.8} &
  \multicolumn{1}{c|}{\hd{9.6}} &
  \ha{85.0} &
  \ha{88.5} \\
\multicolumn{1}{l|}{Apertus-8B} &
  \hmm{20.9} &
  \multicolumn{1}{c|}{\hmm{20.3}} &
  \hd{15.8} &
  \multicolumn{1}{c|}{\hd{14.7}} &
  \ha{59.1} &
  \ha{61.0} \\
\multicolumn{1}{l|}{Apertus-70B} &
  \hmm{17.7} &
  \multicolumn{1}{c|}{\hmm{16.1}} &
  \hd{12.1} &
  \multicolumn{1}{c|}{\hd{8.6}} &
  \ha{82.1} &
  \ha{86.3} \\
\multicolumn{1}{l|}{AMALIA-9B} &
  \hmm{87.5} &
  \multicolumn{1}{c|}{\hmm{91.2}} &
  \hd{33.8} &
  \multicolumn{1}{c|}{\hd{34.8}} &
  \ha{91.8} &
  \ha{95.9} \\ \midrule
\multicolumn{1}{l|}{\textit{\textbf{Open Weight}}} &
   &
  \multicolumn{1}{c|}{} &
   &
  \multicolumn{1}{c|}{} &
   &
   \\
\multicolumn{1}{l|}{Llama-3.1-8B} &
  \hmm{14.1} &
  \multicolumn{1}{c|}{\hmm{12.6}} &
  \hd{10.8} &
  \multicolumn{1}{c|}{\hd{10.6}} &
  \ha{47.8} &
  \ha{47.7} \\
\multicolumn{1}{l|}{Llama-3.3-70B} &
  \hmm{16.6} &
  \multicolumn{1}{c|}{\hmm{15.3}} &
  \hd{8.8} &
  \multicolumn{1}{c|}{\hd{6.7}} &
  \ha{69.5} &
  \ha{67.8} \\
\multicolumn{1}{l|}{Gervásio-8B-ptpt} &
  \hmm{14.9} &
  \multicolumn{1}{c|}{\hmm{12.7}} &
  \hd{12.4} &
  \multicolumn{1}{c|}{\hd{11.1}} &
  \ha{46.3} &
  \ha{46.6} \\
\multicolumn{1}{l|}{Gervásio-70B-ptpt} &
  \hmm{23.3} &
  \multicolumn{1}{c|}{\hmm{22.8}} &
  \hd{12.0} &
  \multicolumn{1}{c|}{\hd{10.8}} &
  \ha{76.4} &
  \ha{75.2} \\
\multicolumn{1}{l|}{Ministral-3-14B} &
  \hmm{19.6} &
  \multicolumn{1}{c|}{\hmm{17.8}} &
  \hd{4.2} &
  \multicolumn{1}{c|}{\hd{4.2}} &
  \ha{93.5} &
  \ha{90.7} \\
\multicolumn{1}{l|}{Qwen3-8B} &
  \hmm{12.0} &
  \multicolumn{1}{c|}{\hmm{10.4}} &
  \hd{9.2} &
  \multicolumn{1}{c|}{\hd{7.9}} &
  \ha{33.2} &
  \ha{32.7} \\
\multicolumn{1}{l|}{Qwen3.5-9B} &
  \hmm{36.1} &
  \multicolumn{1}{c|}{\hmm{35.1}} &
  \hd{13.3} &
  \multicolumn{1}{c|}{\hd{12.2}} &
  \ha{89.1} &
  \ha{86.1} \\
\multicolumn{1}{l|}{Qwen3.5-27B} &
  \hmm{44.4} &
  \multicolumn{1}{c|}{\hmm{44.5}} &
  \hd{12.4} &
  \multicolumn{1}{c|}{\hd{11.9}} &
  \ha{95.1} &
  \ha{93.3} \\
\multicolumn{1}{l|}{Gemma-3-12b} &
  \hmm{25.6} &
  \multicolumn{1}{c|}{\hmm{23.4}} &
  \hd{6.8} &
  \multicolumn{1}{c|}{\hd{4.9}} &
  \ha{93.1} &
  \ha{92.9} \\
\multicolumn{1}{l|}{Gemma-4-E4B} &
  \hmm{23.1} &
  \multicolumn{1}{c|}{\hmm{21.9}} &
  \hd{6.2} &
  \multicolumn{1}{c|}{\hd{4.2}} &
  \ha{92.9} &
  \ha{94.4} \\
\multicolumn{1}{l|}{Gemma-4-31B} &
  \hmm{48.4} &
  \multicolumn{1}{c|}{\hmm{49.2}} &
  \hd{8.9} &
  \multicolumn{1}{c|}{\hd{7.3}} &
  \ha{95.8} &
  \ha{98.0} \\ \midrule
\multicolumn{1}{l|}{\textit{\textbf{Close Source}}} &
   &
  \multicolumn{1}{c|}{} &
   &
  \multicolumn{1}{c|}{} &
   &
   \\
\multicolumn{1}{l|}{Sabiá-4} &
  \hmm{9.4} &
  \multicolumn{1}{c|}{\hmm{8.1}} &
  \hd{7.9} &
  \multicolumn{1}{c|}{\hd{4.6}} &
  \ha{86.9} &
  \ha{87.2} \\
\multicolumn{1}{l|}{Gemini-3-Flash} &
  \hmm{61.2} &
  \hmm{63.7} &
  \hd{4.7} &
  \hd{4.1} &
  \ha{98.7} &
  \ha{99.8} \\ \bottomrule
\end{tabular}%
}
\caption{LLM-as-a-Judge comparison between Gemma-4-31B and Gemini-3-Flash across models and prompting strategies in \dataset. Scores are normalized to a 0--100 scale, where 0 corresponds to pt-BR and 100 to pt-PT, measuring alignment with each Portuguese variant.}
\label{tab_results_llms}
\end{table}

\end{document}